\documentclass[preprint,12pt,authoryear]{elsarticle}

\usepackage{amssymb}

\usepackage{amsmath}
\usepackage{natbib}
\setcitestyle{numbers,square}
\journal{}

\usepackage{hyperref} 
\hypersetup{colorlinks, bookmarks, unicode} 
\usepackage{multicol}

\begin{document}

\begin{frontmatter}



\title{\textbf{Self-Supervised Texture Image Anomaly Detection By Fusing Normalizing Flow and Dictionary Learning}}


\author{ \sffamily Yaohua Guo$^1$,  \sffamily Lijuan Song$^1$$^,$$^2$$^*$, \sffamily Zirui Ma$^1$$^,$$^2$}

\address{	{\sffamily\small $^1$ School of Information Engineering, Ningxia University, Yinchuan 750021; }\\
	{\sffamily\small $^2$ Collaborative Innovation Center for Ningxia Big Data and Artificial Intelligence}\\ {\sffamily\small Co-founded by Ningxia Municipality and Ministry of Education,Ningxia University,Yinchuan 750021 }}

\begin{abstract}
A common study area in anomaly identification is industrial images anomaly detection based on texture background. The interference of texture images and the minuteness of texture anomalies are the main reasons why many existing models fail to detect anomalies. We propose a strategy for anomaly detection that combines dictionary learning and normalizing flow based on the aforementioned questions. The two-stage anomaly detection approach already in use is enhanced by our method. In order to improve baseline method, this research add normalizing flow in representation learning and combines deep learning and dictionary learning. Improved algorithms have exceeded 95$\%$ detection accuracy on all MVTec AD texture type data after experimental validation. It shows strong robustness. The baseline method's detection accuracy for the Carpet data was 67.9$\%$. The article was upgraded, raising the detection accuracy to 99.7$\%$.\\
\end{abstract}


\end{frontmatter}







\section{Introduction}With the advancement of science and technology, industrial production quickly adopts modern manufacturing technologies, changing it from mechanization to intelligence. A common field of computer vision research is quality inspection with machine vision in the context of intelligent industrial production.\par 
Surface anomalies may occur in actual industrial production as a result of technological and environmental constraints. Usually, these surface anomalies result in subpar product quality. Machine vision for quality inspection of product surfaces is a great option to stop unqualified products from reaching the market. Labor costs and inspection mistake rates will be decreased by using machine vision for quality inspection. The goal of surface anomaly detection is to build models from normal samples and use those models for certain detection tasks. During the process of the investigation, it was found that lots of the commonly used anomaly detection methods frequently miss anomalies of particular texture kinds. The traits of texture anomalies are to blame for this. Holes, scratches, color contamination, and other texture anomalies are the most common varieties. These anomalies typically have a small, random size. \par 
When collecting test samples,  the background of acquired example images has the same periodic texture. Anomalies may be challenging to spot on periodic texture backgrounds. On the other hand, in the industrial production process, qualified items outnumber faulty products by a wide margin, and more common to find normal samples. There are fewer abnormal samples and they can be of different kinds. Poor model detection can also be caused by an imbalance between abnormal and normal samples. Based on the aforementioned issues, the work presents a novel solution: self-supervised texture anomaly detection utilizing dictionary learning and normalizing flow. Our strategy is an improvement over the two-stage anomaly detection method described in$^[$$^2$$^0$$^]$. Furthermore, the new model exhibits remarkable robustness, with detection accuracy for various texture anomalies exceeding 95$\%$. The research also successfully raises the Carpet's detection accuracy from 67.9$\%$ to 99.7$\%$.\par
\begin{figure*}[htbp]
	\centering
	\includegraphics[scale=0.32]{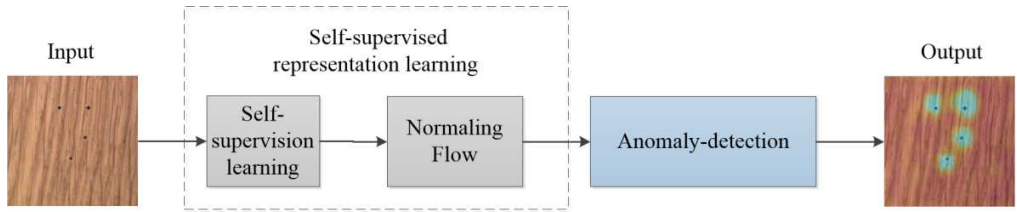}
	\caption{Data flow diagram of the model in this paper}
	\label{figure}
\end{figure*}
\section{Related Work}With the increasing demand for anomaly detection methods based on texture images, many anomaly detection methods for texture images have emerged. These anomaly detection techniques can be divided into three groups depending on whether deep learning is used or not: supervised anomaly detection techniques, unsupervised anomaly detection techniques, and traditional anomaly detection techniques. \\
\textbf{Supervised anomaly detection method.} Supervised anomaly detection method alleviates the problem of anomaly information scarcity by labeling anomalous information on some data. Based on the labeling of the data, supervised anomaly detection methods are classified as supervised anomaly detection, semi-supervised anomaly detection, and weakly supervised anomaly detection. Supervised anomaly detection methods$^[$$^6$$^,$$^7$$^]$ train the model with anomalous samples as negative samples. Such models rely excessively on anomaly information and may overfit known anomalies during training. 
Contrary to the classification task of imbalanced samples$^[$$^8$$^,$$^9$$^,$$^1$$^0$$^]$, the actual detection process of supervised anomaly detection systems may uncover previously undetected anomalies. As a result, supervised anomaly detection should be an open set detection task$^[$$^5$$^]$ that gives the model a potential to find hidden anomalies by studying well-known, apparent anomalies. Semi-supervised anomaly detection$^[$$^2$$^1$$^,$$^3$$^7$$^]$ techniques build a model of normal behavior from normal training samples and employ that model to determine the likelihood of test data is anomalous. By learning from tagged anomaly information, weakly supervised$^[$$^3$$^,$$^4$$^]$ based anomaly detection algorithms enable end-to-end optimization of the anomaly scores. The scoring function of the model is immediately trained using optimized anomaly scores, resulting in significantly higher anomaly scores for anomalous photos than for normal images.\\
\textbf{Unsupervised anomaly detection method.} There are three main categories of unsupervised learning anomaly detection methods: autoencoders, generative adversarial networks and self-supervised learning. Generative adversarial networks$^[$$^1$$^4$$^,$$ ^1$$^5$$^,$$^1$$^6$$^]$ and autoencoders$^[$$^1$$^1$$^,$$^1$$^2$$^,$$^1$$^3$$^]$ will reconstruct the image. The error generated by the reconstructed image is used to determine whether the anomaly exists. Unlabeled data is learned using self-supervised learning anomaly detection$^[$$^1$$^7$$^,$$^1$$^8$$^,$$^1$$^9$$^]$ methods, which provide pseudo-labels to identify anomalies in the data.\\
\textbf{Traditional methods.}
Dictionary learning$^[$$^2$$^2$$^,$$^ 2$$^3$$^,$$^2$$^4$$^]$ is a powerful anomaly detection model that has applications in image restoration$^[$$^1$$^]$, image classification$^[$$^2$$^]$, and image segmentation. It reconstructs an image using a feature dictionary, compares the reconstructed image to the target image, and then determines whether the target image contains anomalies. The overfitting issue brought on by sample imbalance is successfully avoided by the aforementioned technique.\par 
In models that combine deep learning techniques with conventional approaches, neural networks are frequently utilized as feature extractors and traditional approaches are used for anomaly detection. After feature extraction from the input image by a trained neural network in$^[$$^2$$^5$$^,$$^ 2$$^6$$^,$$^2$$^7$$^]$. Even though feature extraction using neural networks produces better results than classic feature extraction techniques, feature extraction on industrial images may not always be applicable to neural networks that have been pre-trained on natural images. Methods$^[$$^2$$^8$$^,$$ ^2$$^9$$^]$ use target datasets to pre-train neural networks for feature extraction, and then encode the extracted features using sparse coding to overcome these problems.\par
The essay decouples two components of representation learning and anomaly detection based on the two-stage images classification$^[$$^3$$^0$$^]$ model and offers several optimizations for each of the two aforementioned parts. It also integrates neural networks with dictionary learning$^[$$^3$$^1$$^]$. Our new approach combines self-supervised learning with dictionary learning while also adding normalizing flow. Both optimization raise the model's detection precision.\par
\section{Motheds}
The self-supervised texture image anomaly detection approach contains two parts: representation learning and anomaly detection. Dictionary learning and anomaly localization are used in the phase of anomaly detection, while CNNs are used in the phase of representation learning to extract features. Self-supervised learning is used in feature extraction to extract the representational properties of the data, which helps to increase the model's ability to extract features, and the inclusion of self-supervised learning in the normalizing flow helps the model understand the subtle differences. As a downstream task of the model, dictionary learning will learn the feature distribution and create a feature dictionary. Recreate the image using the feature dictionary. In order to determine whether the original image is abnormal, it is necessary to compare the differences between the original image and the reconstructed image. Figure 2 shows the architecture of the model.
\begin{figure*}[ht]
	\centering
	\includegraphics[scale=0.21]{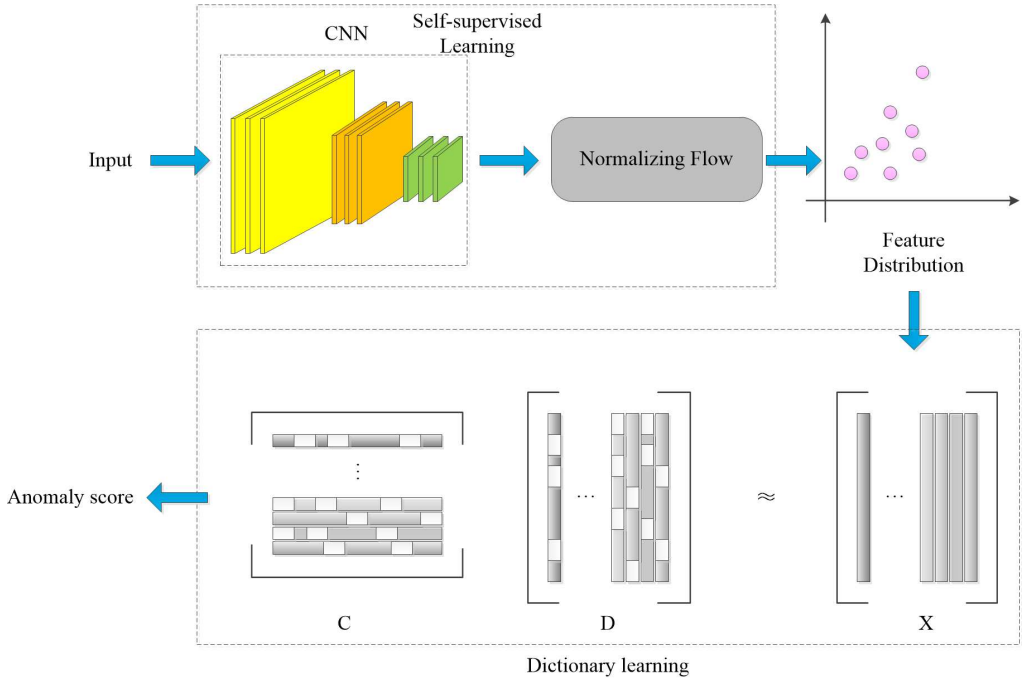}
	\caption{The architecture of  model is shown in the figure above. Extracted data features are used for dictionary learning training. Locate abnormal areas by calculating abnormal scores.}
	\label{figure}
\end{figure*}
\begin{figure*}[ht]
	\centering
	\includegraphics[scale=0.18]{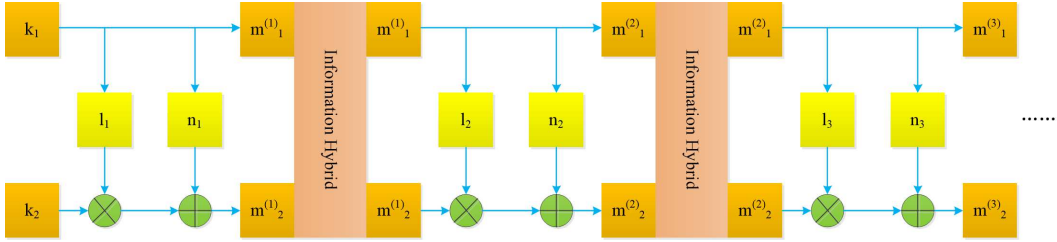}
	\caption{Imitation coupling layer structure diagram}
	\label{figure}
\end{figure*}
\subsection{Representation Learning}
A backbone network will be utilized by deep learning-based anomaly detection metheds to extract features. The backbone network has significant feature extraction capabilities after pre-training on enormous labeled datasets. Real anomaly detection tasks lack a sizable amount of labeled data for backbone network training, and the cost of labeling huge datasets is quite expensive. We try to solve this problem by the model itself, so self-supervised learning is very suitable. By establish auxiliary tasks, self-supervised learning will extract supervised information from unsupervised data. The backbone network is trained using the extracted supervised information to increase the feature extraction capability of the network. For better feature extraction by self-supervised learning, a suitable auxiliary task is essential. A frequent additional task in semantic single classification research is rotation prediction. During the research, texture images with periodic backgrounds lacked subjects, resulting in the rotation prediction task not producing good performance. CutPaste$^[$$^2$$^0$$^]$ is utilized as a auxiliary task in this essay. CutPaste crop a random rectangular patch from the training image and pastes it at any location on the image. With this method, irregularity are generated in the training samples to more realistically simulate realistic anomalies.\par
\subsection{Normalizing Flow}
Anomalies in the texture are typically tiny. By using normalizing flow$^[$$^3$$^3$$^]$, one may learn the distribution of data characteristics and estimate their densities with accuracy. Therefore normalizing flow is sensitive to minute irregularities. In order to improve the model's ability to detect minute anomalies, we attempt to add normalizing flow into feature learning.\par 
Fit unknown complicated feature distributions by normalizing flow. To simulate the unkown distribution, the approach applies a number of bijective transformations to a basic distribution.
\noindent The normalizing flow is defined as follow: 
\begin{equation}
	p_x(x)=p_T(t)|detJ_F(t)|^{-1}, t=F^{-1}(x)
\end{equation}
For $p$ is an initial simple distribution and $x$ obeys $p$ in Equation 1, $F$ stands for the invertible transformation, and both $x$ and $t$ have the same dimension, being D-dimensional.
\noindent The Jacobian matrix in Equation 1 is:
\begin{equation}
	J_F(t) = \left[
	\begin{array}{ccc}
		\frac{\partial F_1}{\partial t_1} & {...} & \frac{\partial F_1}{\partial t_D} \\
		{...} & {...} & {...} \\
		\frac{\partial F_D}{\partial t_1} & {...} & \frac{\partial F_D}{\partial t_D}
	\end{array}
	\right]
\end{equation}\\
A Jacobi matrix is produced by the bijective transformation process on the simple distribution as shown in Equation 2.\par 
We make the assumption that the features in latent space follow some sort of normal distribution during the bijective transformation procedure on feature space and latent space. Figure 3 illustrates the addition of the affine coupling layer structure in Real-NVP to the model.\par
\begin{figure*}[ht]
	\centering
	\includegraphics[scale=0.3]{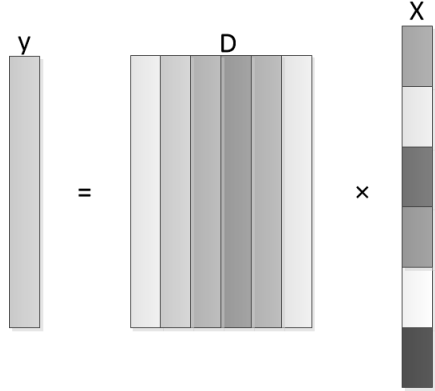}
	\caption{Sparse coding diagram}
	\label{figure}
\end{figure*}
In figure 3, $l$ and $n$ are arbitrary differentiable functions. The partial residual vector of $k_1$ couples additively and multiplicatively with vector $k_2$ to obtain $m^{(1)}_2$:\\
\begin{equation}
	m^{(1)}_2 = l(k_1)\otimes e^{k_2} \oplus n(k_1)k_1
\end{equation}
The vector $m^{(1)}_1$ is the same as vector $k_1$:\par
\begin{equation}
	m^{(1)}_1 = k_1
\end{equation}
New vectors $m^{(1)}_1$ and $m^{(1)}_2$ are spliced again to form a whole. Random ordering and repartitioning of the spliced out new vectors. This method fully mixes information extracted from image. Above operations are repeated several times to form a affine coupling layer. During the multiplicative coupling operation, the exponential function is used to ensure that coefficients are non-zero.\par
Following, the model works various data interpolation processes. On the data, the feature extractor extracts features. A perfect parameter is discovered by the normalizing flow during training. This variable maximize the similarity between the known and unknown distributions:
\begin{equation}
	maximize \quad p_x{(x)}=p_T(t)|detJ_F(t)|^{-1}
\end{equation}
\begin{figure*}[tb]
	\centering
	\includegraphics[scale=0.23]{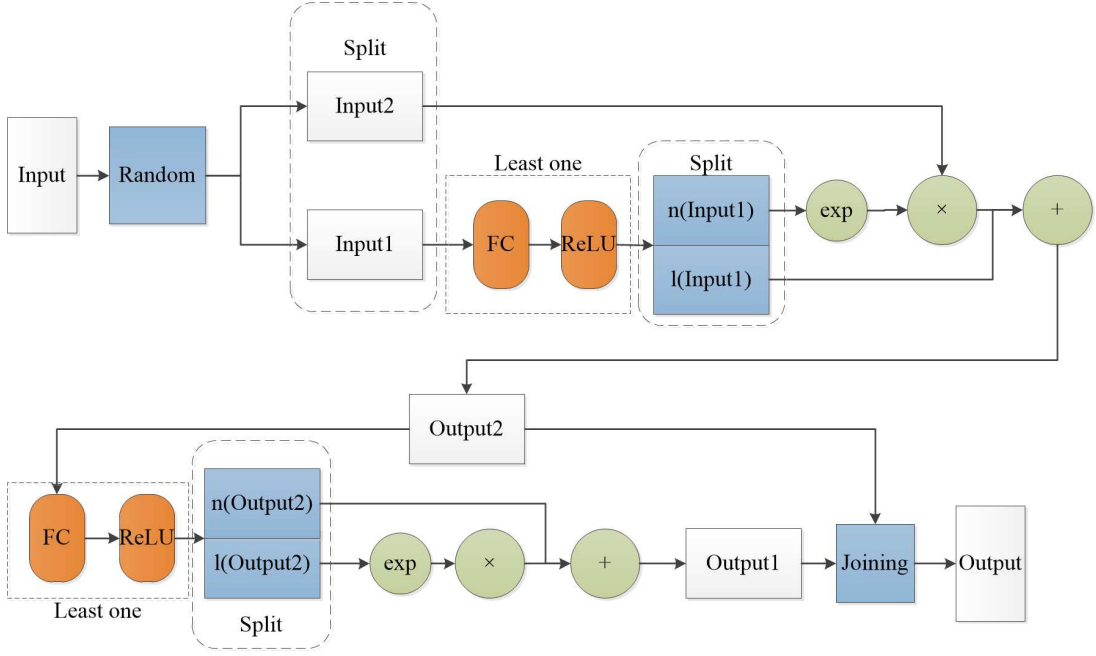}
	\caption{Normaling flow structure diagram}
	\label{figure}
\end{figure*}
The probability function in Equation 5 is applied for determining the parameters. The parameter values are derived by discretizing the continuous function into a discrete function and deriving the derivative, taking the logarithm of likelihood function. Normalizing flow, in contrast to other generative model training methods, trains generative networks just in mapping space, without making distinctions or making approximations about the accuracy of the data.\par
\begin{equation}
	L(x)=\frac{||t||^2_2}{2} - log|det \frac {\partial t}{\partial x}|
\end{equation}
We design a loss function for normalizing flow training as shown in formula six. A typical distribution in the potential space is $t$. mapping the initial distribution to the prospective space. The distribution created after bijective transformation is more similar to the genuine distribution in the potential space, according to a smaller value of the second paradigm based on potential space. The Jacobi matrix's second term in Equation six is a logarithmic determinant, and it represents the scale coefficients' total before the exponential operation. The proper scale of feature transformation must be chosen in order to produce the best outcomes for the model. Part IV of the article will discuss the implementation's specifics. \par
\subsection{Dictionary learning}
Another name for dictionary learning is sparse coding. Its major goal is to extract the crucial characteristics of the data and omit the rest. The repeating arrangement of some important information is ultimately what gives texture image backgrounds their periodicity. This property matches well with the idea of dictionary learning. To improve detection outcomes in the stage of anomaly detection, we attempt to join dictionary learning in the detecting phase.\par
Dictionary learning solves the problem of representation about signal $y$ in N-dimensional space. As shown in Figure 4, The signal $y$ can be expressed as $y \approx Dx$. $y$ is a signal to be reconstructed, $D$ is an overcomplete dictionary, and $x$ denotes the sparse coefficient. The mathematical expression for dictionary learning (sparse coding) can be described as:
\begin{equation}
	\mathop {\min }\limits_x ||x||_0,s.t.||y-Dx||_2 \leq \xi
\end{equation}
$||x||_0$ indicates the number of non-zero vectors in the dictionary. $x$ is a coefficient for constructing the sparse coding and $x \in R^M$, D is an overcomplete dictionary and $D \in R^{N \times M}$, $y$ is an input signal to be reconstructed, $ \xi $ is an arbitrarily small positive number.\par
There are two steps for anomaly detection stage. Firstly, a feature dictionary is used to reconstruct a new image. The error between the reconstructed image and the target image is compared. Whether an image contains anomalies depends on the size of the inaccuracy. In accordance with the aforementioned ideas, selecting a suitable error threshold is essential for anomaly detection, and thresholding the reconstruction error further enhances the model's detection speed.\par
The best sparse coefficients are produced during the optimization iteration process of building a feature dictionary:
\begin{equation}
		 \mathop {\min }\limits_{D,X} \sum_{k=1}^{K}(||y_k-Dx_i||_2^2 + \beta||x_i||_1), 
		 s.t.\forall n||d_n||_2=1
\end{equation}
In the equation eight, dictionary matrix $D={d_1,d_2,... ,d_N}$, where $d_n$ is a feature vector in the dictionary matrix. $x_i \in R_N$ is the sparsity coefficient. The $K$ patches of features $y$ are used as training data for dictionary learning. The second paradigm of the feature vectors in the dictionary must be 1, in order to prevent nonsense values from being produced, due to the uncertainty of multiplying eigenvectors with sparse coefficients.\par
\begin{table}[htbp]
	\centering
	\label{tab:1} 
	\begin{tabular}{cccc ccc}
		\hline\hline\noalign{\smallskip}	
		Category & CutPaste binary(base) & Ours \\
		\noalign{\smallskip}\hline\noalign{\smallskip}
		Carpet & 0.679 & \textbf{0.997}  \\
		Grid & 0.999 & 0.957  \\
		Leather & 0.997 & 0.976  \\
		Tile & 0.959 & \textbf{0.965}  \\
		Wood & 0.949 & \textbf{0.982}  \\
		\noalign{\smallskip}\hline
	\end{tabular}
	\caption{Comparison of AUC values between baseline and modified methods.  The black is the best}
\end{table}
Finding anomalous locations is necessary for an anomaly detection model. To identify image anomaly regions, we employ the method of calculating anomaly scores. The abnormal portions of the anomalous images focus more on high-value anomaly scores. Our model will choose a few regions with the highest anomalous scores to combine, creating a new region that will be the anomaly region located by the model, in accordance with the requirements of the detection task. The five locations with the highest anomalous ratings are combined for the actual detection process. The localized locations have been manually verified to be genuine anomalous regions.
\begin{figure*}[ht]
	\begin{minipage}[t]{0.5\linewidth}
		\centering
		\includegraphics[width=\textwidth]{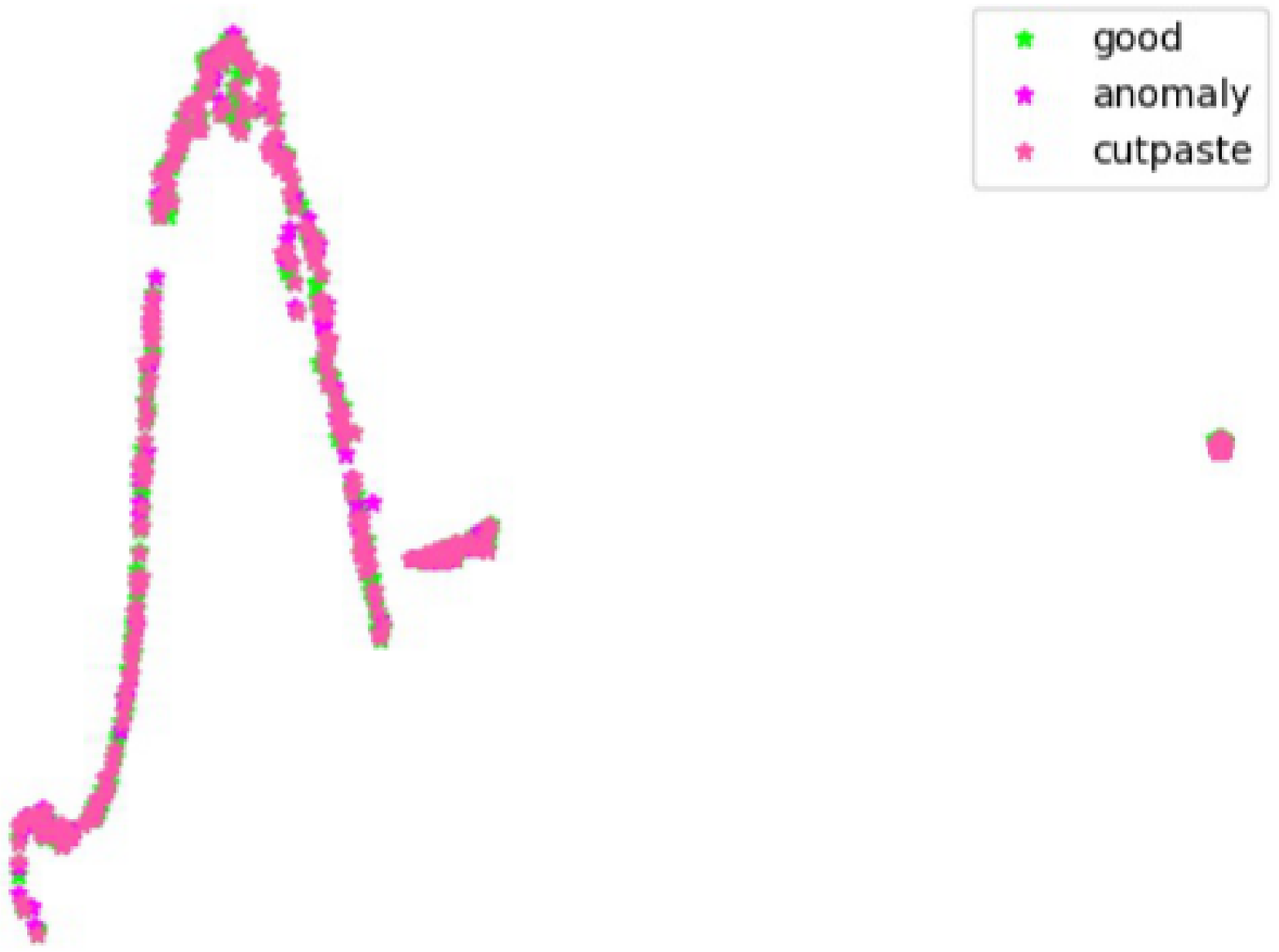}
		\centerline{\small (a) Before join the normalizing flow}
	\end{minipage}%
	\begin{minipage}[t]{0.5\linewidth}
		\centering
		\includegraphics[width=\textwidth]{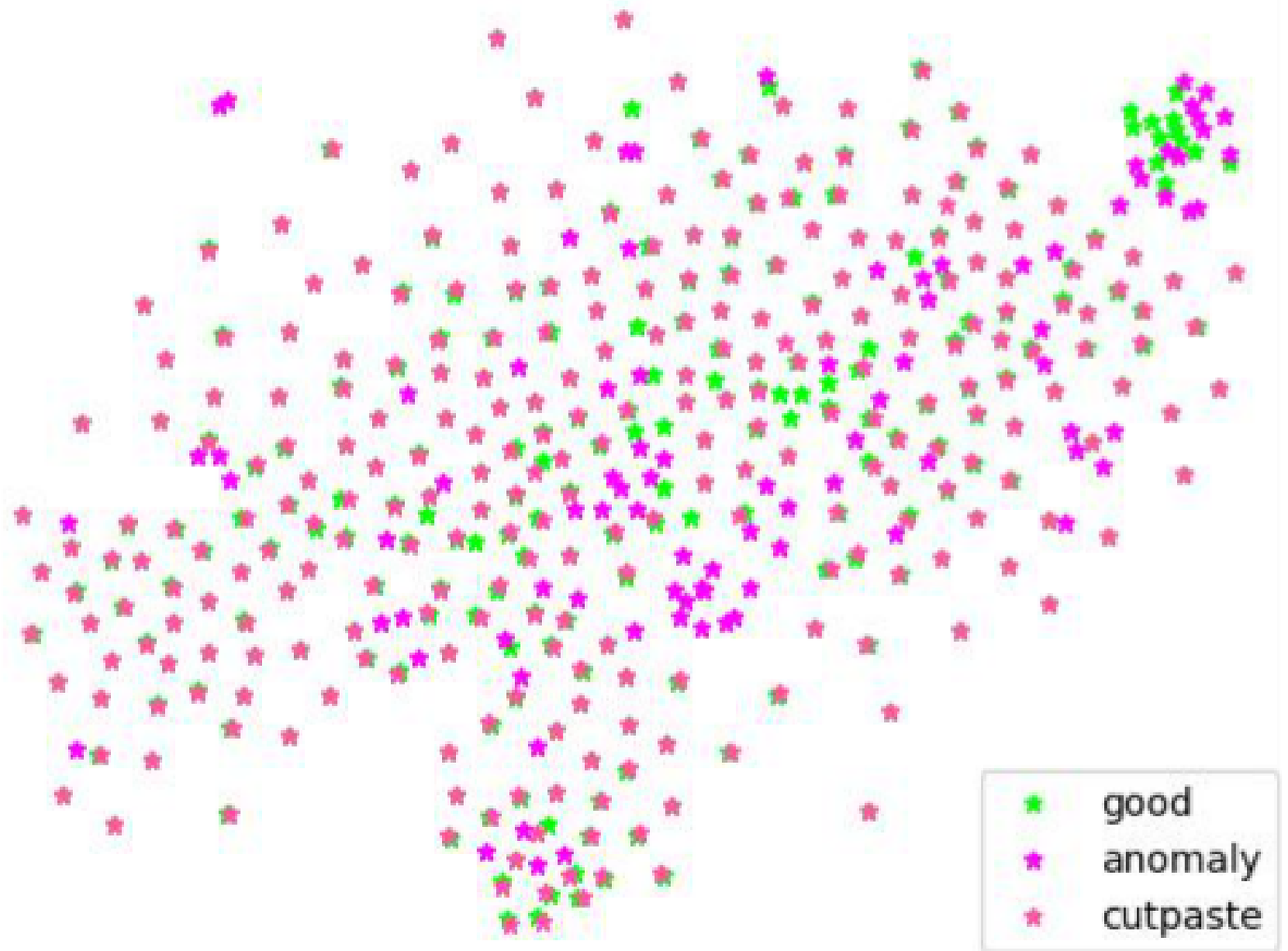}
		\centerline{\small (b) After join the normalizing flow}
	\end{minipage}
	\caption{t-SNE plot of the model on Capet data}
\end{figure*}
\section{Experimental analysis}
\subsection{Datasets}
The MVTec AD dataset, a popular dataset for assessing the performance of unsupervised anomaly detection systems, is used to evaluate the effectiveness of our method. It uses actual industrial anomaly images for its images. 5354 color high-resolution photos totaling more than 70 different anomaly categories are included. In this study, we emphasize the identification of texture anomalies and confirm the model's accuracy in identifying certain texture types, such carpet.The rationale for focusing on texture anomalies in this research is given below\par
In industrial inspection, texture-based industrial anomaly detection is an important detection method. Taking carpet as an example, surface anomalies of these products are often generated by cuts and scrapes, which are less intense and cannot be detected on the product using intensity attributes, and require texture analysis of the sample.Texture is a visual representation of the repetitive patterns of features on the surface of an object. The textures contained in texture images are often complex, and it is difficult for humans to analyze and recognize texture features by eye only. Traditional anomaly detection methods rely on manually selected feature points, which cannot be applied to such texture anomaly data. The use of CNN for image texture learning can avoid the interference generated by textures.\par
Texture images are very common in the field of anomaly detection, and many existing detection models for texture images are not robust enough to be widely applicable to different types of texture images. In the following sections, we hope to build a more robust anomaly detection model for texture images, and we describe the implementation process.
\subsection{Implementation details}
The method, which draws inspiration from the two-stage image classification model$^[$$^3$$^0$$^]$, also separates feature learning and anomaly detection.Based on the baseline model, we optimized the baseline model. There are so two training processes in the improved approach. An image patch is randomly cropped to the training sample and pasted to a randomly chosen area of the sample prior to training. The aforementioned tasks attempt to the greatest extent to imitate abnormalities in actual anomalies. For extracting picture features, our model uses a pre-trained version of AlexNet. Five convolutional layers and three fully connected layers make up AlexNet$^[$$^3$$^5$$^]$, which uses ReLU as its activation function. The AlexNet network structure is less complex than VGG and ResNet's. It can make a model less complex. AlexNet does not perform worse than VGG and ResNet in terms of feature extraction.\par
Normalizing flow helps better learning of minute variations, as was already indicated. Figure 5 illustrates the normalizing flow structure we employ in this paper. The input data are separated into two pieces and sorted at random. On both sections of the input, fully connected activities are carried out concurrently. The internal operations of the fully linked neural network are shown in figure 5 as n and l. Normalizing flow has eight coupling components.
\begin{figure*}[tb]
	\centering
	\includegraphics[scale=0.38]{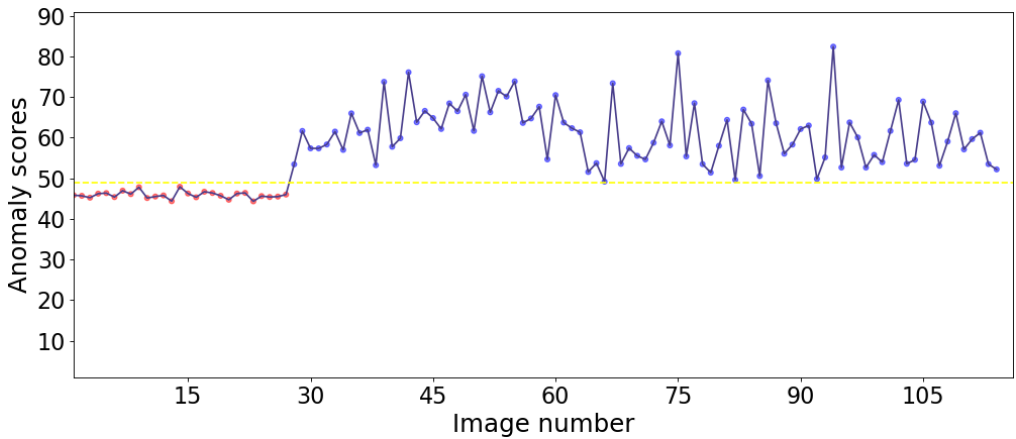}
	\caption{Anomaly scores line graph}
	\label{figure}
\end{figure*}
\begin{table*}[htbp]
	\centering
	\label{tab:2}
	\resizebox{\linewidth}{!}{  
	\begin{tabular}{cccc cccc c}
		\hline\hline\noalign{\smallskip}	
		Category & DSEBM$^[$$^3$$^9$$^]$ & GANomaly & 1-NN & OC-SVM$^[$$^3$$^8$$^]$ & Sparse coding & P-SVDD$^[$$^3$$^7$$^]$ & Ours \\
		\noalign{\smallskip}\hline\noalign{\smallskip}
		Carpet & 0.437 & 0.699 & 0.811 & 0.627 & 0.58 & 0.929 & \textbf{0.997} \\
		Grid & 0.619 & 0.708  & 0.557 & 0.410 & 0.89 & 0.946 & \textbf{0.957}\\
		Leather & 0.841 & 0.842  & 0.903 & 0.880 & 0.95 & 0.909 & \textbf{0.976}\\
		Tile & 0.690 & 0.834  & \textbf{0.994} & 0.876 & 0.86 & 0.978 & 0.965\\
		Wood & 0.952 & 0.794  & 0.934& 0.953 & 0.97 & 0.965 & \textbf{0.982}\\
		\noalign{\smallskip}\hline
	\end{tabular}
	}
	\caption{Comparison with AUC values of other methods}
\end{table*}
\subsection{Experimental results}
The t-SNE plots produced by the model using the Carpet data are displayed in Figures 6(a) and 6(b). Figure 6(b) makes it obvious that the data processed by the auxiliary task and the normal and abnormal data no longer overlap. This shows that normalizing flow enhances the model's capacity to detect minute irregularities. The model is more sensitive to minute variations in the data.\par
The input size for the training phase's photos is 128 and the learning rate is 2. During the experiment, it was discovered that the loss value of the model stabilized after 270 epochs of training. Input images have a batch size of 2. A model created by self-supervised representation learning extracts features from training samples during the anomaly detection phase. There are 32 inputs in the batch. Extracted features will form feature dictionaries. For image reconstruction and sparse coding, feature dictionaries are crucial. 
A judgment on the performance of the anomaly detection model is also crucial for anomaly detection. ROC curves are used to assess model performance. The ROC curve is a series of values obtained under particular circumstances using various evaluation criteria. AUC value is the area of the ROC curve with the coordinate axis. Higher AUC value means better detection accuracy of the model.\par
As indicated in Table 1, during the trial we compared the outcomes of the upgraded approach with the baseline method. The detection accuracy of the modified approach in this work is enhanced by 31$\%$ on Carpet data. Additionally, the revised approach slightly enhances the Tile and Wood data. As demonstrated in Table 2, in addition to comparisons with the baseline approach, the approach also included comparisons with Sparse coding$^[$$^2$$^2$$^]$, 1-NN$^[$$^4$$^0$$^]$, GANomaly$^[$$^3$$^7$$^]$, and other approaches.   \par
In contrast, it is discovered that the enhanced method has significant robustness and does well in each of the five texture datasets. It may be used to identify anomalies in a variety of texture types. The table in bold is the best performance.\par
By computing anomaly scores, abnormal locations are localized. To distinguish between normal and anomalous regions, an error threshold is set. Figure seven shows a line graph of anomaly scores for partial Carpet images. Red dots in the figure represent normal images and blue dots represent anomalous images. From figure seven, we can find that the scores of normal images are significantly lower than those of abnormal images. \par
In summary, the article describe a method for detecting textural anomalies that combines self-supervised learning and dictionary learning. The proposed anomaly detection method's performance is clearly described in Tables 1 and 2. The research backs up the idea that a normalizing flow can aid the model in capturing tiny anomalies more effectively. It improves  the detection accuracy of the model and makes the model more sensitive to minute anomalies. Additionally, neural networks can improve the efficiency of conventional methods for feature extraction. Combining convolutional neural networks with traditional methods will significantly improve the performance of detection methods.\par  
\section{Discussion}
In this work, different strategies are adopted to solve different problems faced by texture anomaly detection, and satisfactory results are achieved. In the face of unlabeled data, the self-supervised learning strategy is used to reduce the training cost of the model. Normalizing flow is used to solve the problem of abnormal texture smallness. In order to meet the needs of real-time detection, dictionary learning is used as the downstream task of the model to improve the detection efficiency and ensure the detection accuracy. In the later research work, we will extend the idea of this paper to other different types of data, and strive to explore more general anomaly detection methods.
\section{Funding}
This work has been partially supported by the National Natural Science Foundation of Ningxia (2019AAC03034) and the Key Research and Development Program of Ningxia (2019BEB04023 and 2021BEE03013).

	



\end{document}